\documentclass[journal, onecolumn, 11pt, draftclsnofoot]{IEEEtran}
\pdfoutput=1
\usepackage{amsmath}

\DeclareMathOperator*{\argmax}{arg\,max}
\usepackage{mdwlist}
\usepackage{pifont}
\usepackage{float}
\usepackage[utf8]{inputenc} 
\usepackage[T1]{fontenc}    
\usepackage{hyperref}       
\usepackage{url}            
\usepackage{booktabs}       
\usepackage{amsfonts}       
\usepackage{nicefrac}       
\usepackage{microtype}      
\usepackage{xcolor}         
\usepackage{comment}
\usepackage{multirow}
\usepackage{graphicx}
\usepackage{tabularx}
\usepackage{setspace}
\usepackage{hyperref}
\hypersetup{
    colorlinks=true,
    urlcolor=blue,
}
\usepackage{amsthm}
\usepackage{csquotes}
\usepackage{soul}

\theoremstyle{definition}
\newtheorem{definition}{Definition}[section]

\begin{document}

\onecolumn 
\begin{basedescript}{\desclabelstyle{\pushlabel}\desclabelwidth{6em}}

\item[\textbf{Citation}]{G. AlRegib and M. Prabhushankar, "Explanatory Paradigms in Neural Networks," \emph{IEEE Signal Processing Magazine, Special Issue on Explainability in Data Science}, accepted on Feb. 18 2022.}

\item[\textbf{Review}]{Data of White Paper Submission : 22 March 2021 \\ Date of Invitation : 29 April 2021 \\ Date of Initial Submission : 25 August 2021 \\ Date of First Revision : 16 November 2021 \\ Date of Second Revision : 18 January 2022 \\ Date of Accept: 18 Feb 2022}

\item[\textbf{Codes}]{\url{https://github.com/olivesgatech/Explanatory-Paradigms}}

\item[\textbf{Copyright}]{\textcopyright 2022 IEEE. Personal use of this material is permitted. Permission from IEEE must be obtained for all other uses, in any current or future media, including reprinting/republishing this material for advertising or promotional purposes,
creating new collective works, for resale or redistribution to servers or lists, or reuse of any copyrighted component
of this work in other works. }

\item[\textbf{Contact}]{\href{mailto:mohit.p@gatech.edu}{mohit.p@gatech.edu}  OR \href{mailto:alregib@gatech.edu}{alregib@gatech.edu}\\ \url{http://ghassanalregib.com/} \\ }
\end{basedescript}
\thispagestyle{empty}
\newpage
\clearpage
\setcounter{page}{1}


%
\title{Explanatory Paradigms in Neural Networks}

\author{Ghassan~AlRegib,~\IEEEmembership{Fellow,~IEEE,}
        and~Mohit~Prabhushankar,~\IEEEmembership{Member,~IEEE}}

\markboth{Signal Processing Magazine, Special Issue on Explainability in Data Science}%
{Prabhushankar \MakeLowercase{\textit{et al.}}: Explanatory Paradigms in Neural Networks}

\maketitle

\begin{abstract}
In this article, we present a leap-forward expansion to the study of explainability in neural networks by considering explanations as answers to abstract reasoning-based questions. With \emph{P} as the prediction from a neural network, these questions are \emph{`Why P?'}, \emph{`What if not P?'}, and \emph{`Why P, rather than Q?'} for a given contrast prediction \emph{Q}. The answers to these questions are observed correlations, observed counterfactuals, and observed contrastive explanations respectively. Together, these explanations constitute the abductive reasoning scheme. We term the three explanatory schemes as observed explanatory paradigms. The term observed refers to the specific case of \emph{post-hoc} explainability, when an explanatory technique explains the decision $P$ after a trained neural network has made the decision $P$. The primary advantage of viewing explanations through the lens of abductive reasoning-based questions is that explanations can be used as reasons while making decisions. The post-hoc field of explainability, that previously only justified decisions, becomes active by being involved in the decision making process and providing limited, but relevant and contextual interventions. The contributions of this article are: ($i$) realizing explanations as reasoning paradigms, ($ii$) providing a probabilistic definition of observed explanations and their completeness, ($iii$) creating a taxonomy for evaluation of explanations, and ($iv$) positioning gradient-based complete explanainability's replicability and reproducibility across multiple applications and data modalities, ($v$) code repositories, publicly available at \url{https://github.com/olivesgatech/Explanatory-Paradigms}
\end{abstract}

\section{Introduction}
\label{sec:Intro}

The generalizability of machine learning has fostered its ubiquitous usage across diverse applications within Artificial Intelligence (AI). AI takes an inferential and interpretive role in safety-critical fields of medicine, remote sensing, and bioinformatics~\cite{freitas2014comprehensible} among others. For example, AI algorithms detect cardiovascular risk factors in humans through retinal fundus images~\cite{poplin2018prediction}. In~\cite{carton2016identifying}, the authors propose an AI-based intervention system to identify early red flags in law enforcement to prevent police-public adverse events. In such sensitive applications, it is apparent that the role of inference in an AI system is not a binary conclusion of whether risks exist. Rather, AI inference must provide an explanation about the risks involved in the decision, thereby allowing humans to assess and act on the risks. Consequently, the authors in~\cite{poplin2018prediction} provide a visual explanatory map to show why and where an AI model detects cardiovasular risk symptoms. A number of works exist that track the larger concept of interpretability in AI. The authors in~\cite{doshi2017towards} provide a comprehensive overview of AI interpretability and its larger denomination with related fields of privacy, transparency, safety, explainability, and ethics. In this article, we specifically focus on explainability on visual data for a specific modality of machine learning, namely neural networks.

The generalization and task-independent modeling capabilities of neural networks have resulted in their widespread applicability. Generalization refers to a neural network's ability to regularize to unseen data. Effective generalization is required for large-scale learning. Task-independent modeling refers to transference of low-level features between tasks and networks. For instance, pretrained neural nets that are trained on natural images are used on computed seismic images~\cite{shafiq2018towards}. These two capabilities coupled with hardware improvements have allowed neural nets to achieve state-of-the-art performance on a number of tasks. Specifically in image classification, neural networks surpassed top-5 human accuracy of $94.9\%$ on ImageNet dataset~\cite{he2016deep}. While gaining such high performance accuracy is an achievement, it is equally important to understand the reasons that lead to such high accuracy rates. Thus, it is imperative to study explainability in neural networks to better understand the underlying science behind the phenomenon. However, explainability in such networks is challenging since neural nets do not conform to the traditional notion of AI interpretability. An example of AI interpretability is through decision trees~\cite{breiman1984classification}. The idea that neural networks should obtain explainability similar to decision trees is not unreasonable. Decision trees are non-linear models that create local binary partitions. This is not dissimilar to neural net neurons with ReLU activation functions that either propagate or suppress features. Providing such an explanation requires a complete understanding of features at each node. While decision trees are designed using disentangled and interpretable features, neural networks learn such features along with feature interactions. This challenge has recently been compounded by the scale of neural networks. The natural language processing model GPT-3~\cite{brown2020language} consists of 175 billion parameters. Parsing all parameters and decisions within such models is impractical. Hence, explainability in neural networks is abstract. This abstract definition of explainability in neural networks is motivated by the broad definition of explanations themselves.

\begin{figure*}[!t]
\begin{center}
\minipage{\textwidth}%
\includegraphics[width=\linewidth]{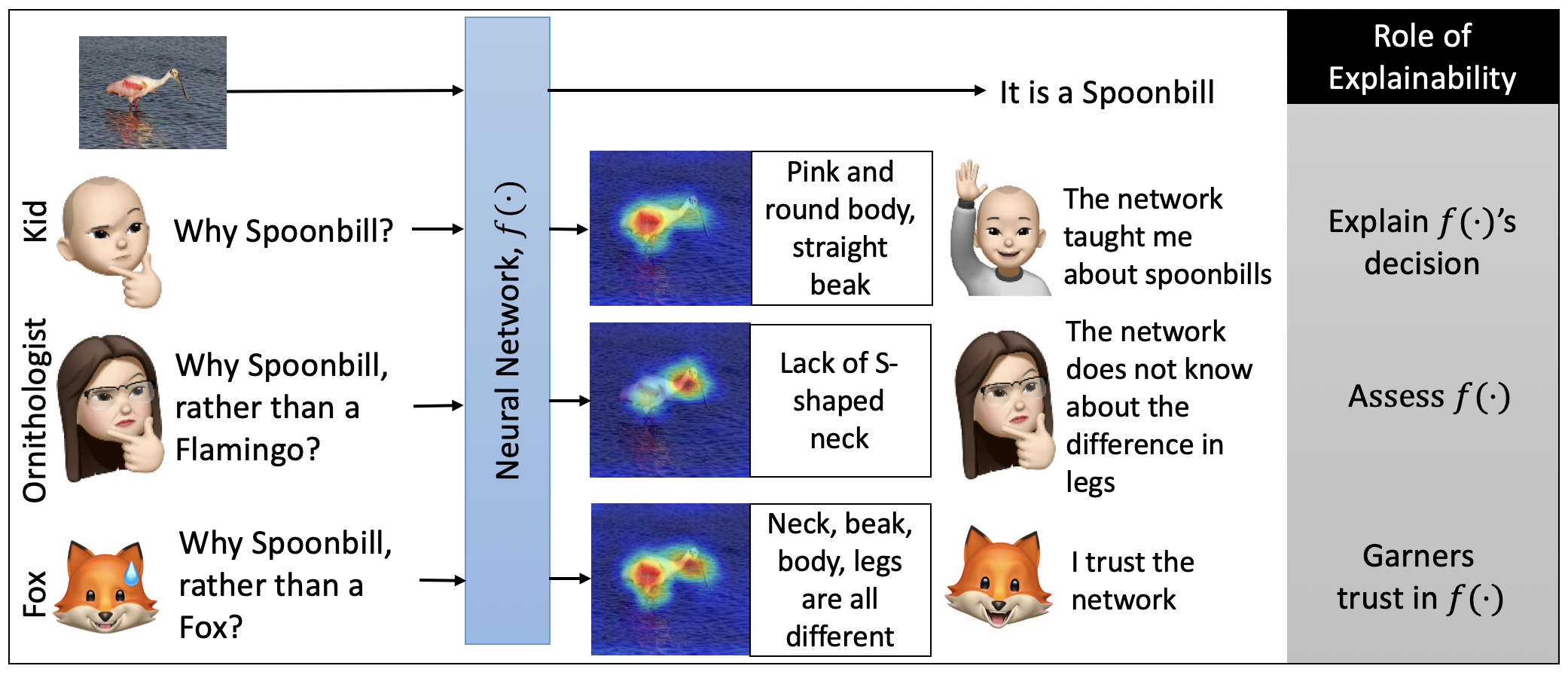}
\endminipage
\vspace{-3mm}
\caption{Role of explainability demonstrated through targeted toy examples. Note that $f(\cdot)$ is not trained to answer abstract questions. The explanations provide additional insights beyond the trained solution.}\vspace{-0.7cm}\label{fig:Intro}
\end{center}
\end{figure*}

Explanations are defined as a set of rationales used to understand the reasons behind a decision~\cite{kitcher1962scientific}. If the decision is based on visual characteristics within the data, the decision-making reasons are visual explanations. Consider the image of a spoonbill in Fig.~\ref{fig:Intro}. A trained neural network correctly recognizes it as a spoonbill. Consider the form of explanations that are acceptable for such a decision. In Fig.~\ref{fig:Intro}, we present three such toy explanations all of which allude to the general goals of explainability. All three explanations are abstract answers to specific questions. When a kid is interested to know \emph{`Why Spoonbill?'}, the network's explanation highlights the body and head of the bird leading to that decision. This is an abstract explanation that requires humans to take the extra step of knowing that the highlighted regions depict the spoonbill's flat-beak and its pink and round body. For the neural network, these characteristics determine the class of the given bird. Note that these are not causal factors. Flamingos have the same body shape and color that the explanation highlights. Consider the scenario when an ornithologist, armed with the knowledge that flamingos and spoonbills share the same body shape and color, asks the question \emph{`Why Spoonbill, rather than a Flamingo?'}. Such a targeted and contextualized question elicits the explanation that the two birds differ in the shape of their necks - specifically, the lack of an S-shaped neck in a spoonbill is highlighted as to why it is not a flamingo. The relevance of the lack of an S-shaped neck is accentuated by the question, \emph{`Why Spoonbill, rather than a Flamingo?'}. Such an explanation can be used to assess the network. In the last toy example, the network answers the question, \emph{`Why Spoonbill, rather than a Fox?'} by highlighting all the body parts of the bird. Note that the network is only trained to classify birds and has \emph{not} explicitly been trained to answer these questions. Hence, all these intelligible and abstract explanations together serve to garner \emph{trust} in the network and its inference capabilities.

The central thesis of this article is to motivate and categorize existing abstract explanatory techniques based on specific questions. We consider three types of questions, namely observed correlations \emph{`Why P?'}, observed counterfactual \emph{`What if?'}, and observed contrastive \emph{`Why P, rather than Q?'}. Here $P$ is any prediction from a neural network, like a spoonbill, and $Q$ is any contrast class, like a flamingo or a fox. The answers to these questions are \emph{observed} in the sense that the decision is already made. Together, these three questions provide complementary explanations that we term as explanatory paradigms. Their complementary nature allows us to define complete explanations as justifications for any and all posed contextually relevant questions. For the observed counterfactual and contrastive paradigms, there is an additional intervention that allows for their respective questions. These explanations are not \emph{interventionist} from a causal perspective~\cite{pearl2009causal}. Interventions refer to manipulations in data that are designed to probe a network's class causality and construct counterfactuals. In this article, the interventions considered are based on a network's notion of classes, through gradients. By considering such interventions, we elicit model behaviors that provide responses to the ornithologist and fox from Fig.~\ref{fig:Intro} without the need for engineered interventions. The exact definition of each of the question-based paradigm and its relationship with existing categorizations is presented in Section~\ref{sec:Paradigms}. We start by describing the existing definitions of explainability in neural networks and provide our own definition that is both intuitive and mathematically rigorous in Section~\ref{sec:Background}. We then summarize the explanatory paradigms and their advantages in Section~\ref{sec:Paradigms} before combining them into complete explanations. A taxonomy of explanatory evaluation is provided in Section~\ref{sec:Evaluation} followed by a summary of existing methods of explaianbility in Table~\ref{tab:Results_Compare}. We discuss existing challenges in explainability that inform future research in Section~\ref{sec:Challenges} before concluding in Section~\ref{sec:Conclusion}.

\section{Background}
\label{sec:Background}
We first discuss requisite neural network preliminaries. While the focus is on classification networks, the preliminary discussion extends to other discriminative and generative networks as well. This is followed by existing and our own definitions of explainability.
\subsection{Neural Network Preliminaries}
\label{subsec:Prelim}
Classification neural networks learn non-linear transformations to obtain discriminative representation spaces on any given data. The inherent mechanisms behind the discriminative process can be formulated as follows. Let $f(\cdot)$ be a neural network trained to distinguish among $N$ classes. If $x$ is any input to the network, the output logits in a classification network are given by
\begin{equation}\label{eq:Network}
\begin{gathered}
    \hat{y} = f(x), \forall  \hat{y} \in \Re^{N\times 1},
\end{gathered}
\end{equation}
\noindent where $\hat{y}$ is an $(N\times 1)$ vector. The predicted class of $x$, given by $P$, is the index of the maxima of $\hat{y}$. i.e.,  
\begin{equation}\label{eq:Filter}
\begin{gathered} 
    P = \argmax \hat{y}, P\in [1,N].
\end{gathered}
\end{equation}

During training, the network is penalized for predicting the wrong class. Penalty is calculated by using an empirical loss function, $J(P, y)$, where $y$ is the true label for input $x$. This loss is backpropagated through the network. The network parameters are updated based on gradients $\frac{\partial J(P, y)}{\partial \theta}$ where $\theta$ are the network parameters, namely its weights and biases. Mathematically, this update is given by, 
\begin{equation}\label{eq:Backprop}
    \theta' = \theta - \frac{\partial J(P, y)}{\partial \theta},
\end{equation}
where $\theta'$ are the parameters after the update. A full description of backpropagation and the associated math is presented in~\cite{rumelhart1986learning}. The gradients can be backpropagated all the way back to the input $x$ to obtain $\frac{\partial J(P, y)}{\partial x}$ or to the activations at some layer $L$, given by $A_L$, to obtain $\frac{\partial J(P, y)}{\partial A_L}$. These quantities represent sensitivity which is defined as a measure of the change in the prediction loss given a small perturbation either in $x$ or $A_L$. Parameter gradients and sensitivity are widely used as explanatory features among a number of explanatory methods including~\cite{selvaraju2017grad, prabhushankar2020contrastive, prabhushankar2021Contrastive, prabhushankarCausal, chattopadhay2018grad, smilkov2017smoothgrad, springenberg2014striving}. 

\subsection{Interventions, causality, and \emph{post-hoc} explanations}
\label{subsec:Interventionist}
Interventions in data are manipulations that are designed to test for causal factors~\cite{pearl2009causal}. The authors in~\cite{scholkopf2021toward} use the presence of interventions to motivate two types of data, namely observational and interventionist. The interventionist data is specifically engineered to test for causal variables. It employs the \emph{do} operator and is generally given as $\mathbb{P}(Y|do(x))$ for data $x$ and label $Y$. The \emph{do} operator represents interventions. However, all possible interventions in data can be long, complex and impractical~\cite{lopez2017discovering}. Even with interventions, it is challenging to estimate if the resulting effect is a consequence of the intervention or due to other uncontrolled interventions. Instead of interventions, \emph{post-hoc} explanations highlight features using a trained network that aid inference. In this article, we use the term \emph{observed} to denote a specific case of \emph{post-hoc} explainability after a network has made its decision. We then construct gradient-based observed interventions to create complete observed explanations.
\subsection{Explanation Maps}
\label{subsec:ExplanationMap_Background}
The definition of explanatory maps in neural networks has evolved over time. The early works for explainability~\cite{zeiler2014visualizing, mahendran2015understanding} consider explanations to be processes that illustrate the workings of neural networks. The deconvolution-based explainability technique proposed in~\cite{zeiler2014visualizing} shows that at different layers and different kernels, different structures in the data activate to create local decisions. Deconvolution, unpooling, and rectification are used to reconstruct and visualize activations at all layers. These visualizations are analyzed based on layers and kernels to obtain insight regarding individual parameters within the network. Such a hierarchial representation is followed in~\cite{zhang2018interpreting} where the authors consider the neural network as an explanatory graph. From this methodology of obtaining explanations, one could define explanations as visualizations that require further deductions regarding the network and its outputs. We term such explanations as \emph{indirect explanations}. An alternative definition of explainability is based on~\cite{zhou2016learning} that defines visual explanations as localization maps. Such a definition is based on the argument that neural networks, with no explicit training, implicitly localize objects that they recognize. We term these types of explanations as \emph{direct explanations}. Direct explanations passively highlight all regions in an image that are used to predict $P$.~\cite{zhou2016learning} produces a Class Activation Map (CAM) that highlights parts of the image leading to the final decision. Direct explanations are passive in the sense that they retroactively justify decisions.

Recently however, direct and indirect definitions of explanations are insufficient to categorize some of the newer explanatory techniques~\cite{selvaraju2017grad, prabhushankar2020contrastive}. The authors in~\cite{selvaraju2017grad} show explanations for any class in any given image even when objects from that class are not present in the image. The authors in~\cite{prabhushankar2020contrastive} show the contrast between any two classes in the image. This is depicted in the contrast between a spoonbill and flamingo in Fig.~\ref{fig:Intro}. Neither the definitions of localization, nor reconstruction encompass these works. However, such works lead to explanations taking an active role in neural nets by participating in decision making or assessing their parent networks~\cite{prabhushankar2021Contrastive}. These explanations encompass all three explanatory roles sought in Fig.~\ref{fig:Intro}. In this article we propose a new and simple probabilistic definition for explanations that combine newer and older explainability techniques, and construct a more comprehensive paradigm.

In order to better understand and distinguish the definitions of explainability, consider an image $x$ fed into a trained network $f(\cdot)$. The network is trained to distinguish $x$ between $N$ classes as $P \in Y$. Let $\mathcal{T}$ be the set of all features on which the network has learned to base its decision. In other words, a network infers that $x \in P$, if $x$ has features $\mathcal{T}_p$. Hence $\mathcal{T} = [\mathcal{T}_1, \mathcal{T}_2 \dots \mathcal{T}_M]$, where $M$ is the total number of features that is learned by a network. Note that we are not considering $\mathcal{T}_i, i\in [1,M]$ to be the causal features. In other words, $\mathcal{T}_i \cap \mathcal{T}_j \neq {0}, \forall i \neq j$. Features from multiple sets constitute $\mathcal{T}_p$ and are used to make inferences on $x$, i.e, $\mathcal{T}_p = \bigcup\limits_{i=1}^{K_p} \mathcal{T}_i$, $K_p \in [1,M]$. We define an explanatory map $\mathcal{M}$ as follows:

\begin{definition}[Observed Explanation]\label{def:explanation}
An explanatory map is any function $\mathcal{M}(\cdot)$ that maximizes the conditional probability $\mathbb{P}(\bigcup\limits_{i=1}^{K} \mathcal{T}_i|Y)$, where $Y$ is any possible class, $K \in [1,M]$ and $\bigcup\limits_{i=1}^{K} \mathcal{T}_i$ is the union of all required features $\mathcal{T}_i$, that specifically lead to $Y$.
\end{definition}

In Definition~\ref{def:explanation}, we assume the decision $Y$ before maximizing the features $\mathcal{T}_i$. However, by not specifying the value for $Y$, we allow for interventions in labels to create contrastive paradigms. By specifying the features as non causal $\mathcal{T}_i$, we open the possibility of changing the features to create counterfactual paradigms. This allows specific interventions in $Y$ and $\mathcal{T}_i$. Definition~\ref{def:explanation} is not \emph{ causal interventionist} since we neither alter the feature set $\emph{T}$ - interventionist counterfactual - nor probe on $x$ for interventionist causality. Our definition provides for restricted interventions for a decision $P$ and we term it \emph{observed}. Our observed definition of explanation comprises both indirect and direct explanatory techniques. Direct and indirect explanations maximimally highlight features that explain the decision $P$. In other words, their goal is to maximize $\mathbb{P}(\mathcal{T}_p|P)$ for the specific instance of $Y=P$. Our general definition explains any decision, $Y$, by maximally highlighting features $\bigcup\limits_{i=1}^{K} \mathcal{T}_i$. The newer explanatory techniques like~\cite{selvaraju2017grad, prabhushankar2020contrastive} that do not conform to direct and indirect definitions of explanations can be explained using Definition~\ref{def:explanation}. This is presented in Section~\ref{sec:Paradigms}. 

\begin{figure*}[!t]
\begin{center}
\minipage{\textwidth}%
\includegraphics[width=\linewidth]{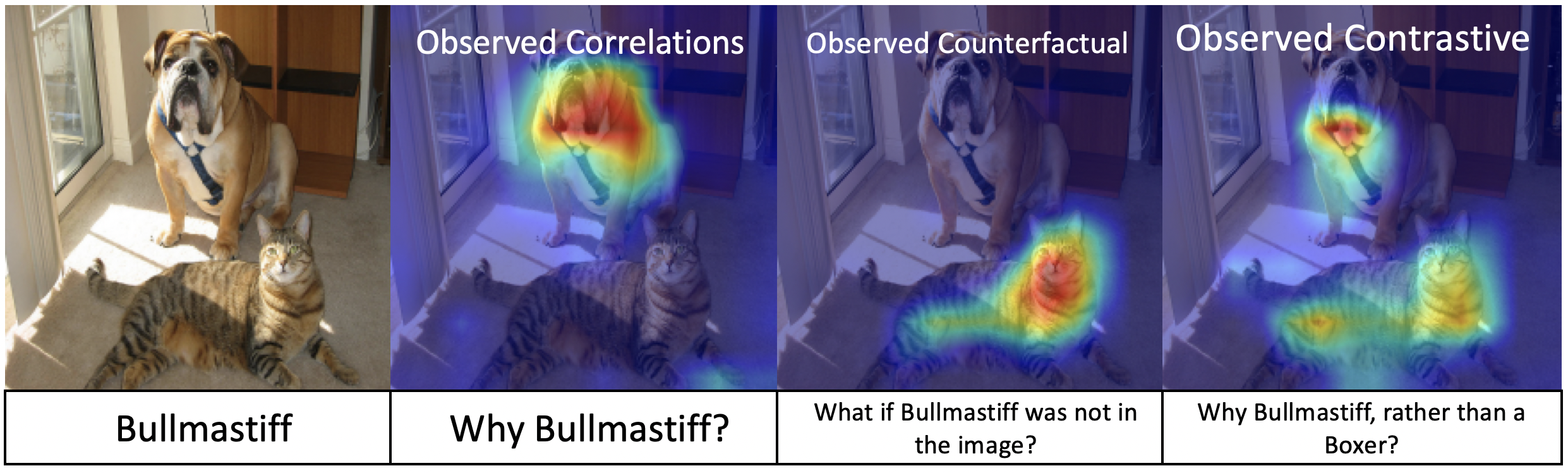}
\endminipage
\vspace{-3mm}
\caption{Three paradigms of explanations, each answering observed correlation, counterfactual, and contrastive questions, respectively.}\vspace{-0.7cm}\label{fig:Paradigms}
\end{center}
\end{figure*}

\section{Explanatory Paradigms}
\label{sec:Paradigms}
\subsection{Observed Explanatory Paradigms via Reasoning}
\label{subsec:Reasoning}
We first motivate our observed paradigms through reasoning. The primary goal of neural networks is to infer decisions. Any network $f(\cdot)$, is trained to maximize $\mathbb{P}(P|\mathcal{T}_p)$, i.e, $\mathcal{T}_p$ is the reason that leads $f(\cdot)$ to predict $P$. Hence, explanations are a means to extract these reasons. There are three ways in which reasoning can occur - deductive, inductive, and abductive reasoning~\cite{prabhushankar2021Contrastive}. Deductive reasoning occurs when networks are provided with a set of features and logic to use the given features. Since both the features, and the logic to manipulate the features are learned by the network, deductive reasoning does not play a large role in neural nets. Inductive reasoning forms the core of the reasoning behind current networks. A set of parameters are learned consisting of some underlying rules and knowledge. During inference, neural nets associate given data with the knowledge in their parameters and make their predictions with a probability. Explanations are then generated to justify such a decision. Direct and indirect explanations are a product of this reasoning scheme. The third form of reasoning is abductive reasoning. The primary difference between inductive and abductive reasoning is the usage of explanations. In abductive reasoning, explanation is any hypothesis that supports a prediction. In Fig.~\ref{fig:Intro}, all three explanations support the prediction that the given image is that of a spoonbill. The network can detect the presence of the pink and round body, and straight beak to inductively infer the class of the bird as spoonbill. Or, alternatively, it can hypothesize as to why the bird cannot be a flamingo - lack of S-shaped neck - and then decide that the bird has to be a spoonbill. Such an inference is abductive. A number of such hypotheses can be constructed, fused and incorporated together, and then used to make a decision. 

In this article, we present three specific ways of generating such hypotheses - as answers to causal questions \emph{`Why P?'}, counterfactual questions \emph{`What if not P?'}, and contrastive questions \emph{`Why P, rather than Q?'}. However, since we use observed techniques, the network provides correlation features to the causal question and we name it observed correlation. In Section~\ref{subsec:Complete}, we show that these three explanatory paradigms are sufficient to generate probabilistically complete explanations. A visual depiction of the three paradigms is presented in Fig.~\ref{fig:Paradigms}. Consider a bullmastiff as the given input image. The observed correlation explanation answers \emph{`Why Bullmastiff?'} by highlighting the face of the dog.\footnote{Grad-CAM~\cite{selvaraju2017grad} is used to obtain this explanation.} The question \emph{`What if the bullmastiff was not in the image?'} is a counterfactual question. As highlighted using the technique in~\cite{selvaraju2017grad}, the network would have made a decision based on the cat for the given counterfactual. The contrastive question is of the form \emph{`Why Bullmastiff, rather than Boxer?'}. The heavy jowls of the bullmastiff and the presence of a cat preclude the network from predicting a boxer as highlighted by Contrast-CAM from~\cite{prabhushankar2020contrastive}. Contrastive explanations take into account the context provided by the questionnaire thus making explanations more relevant. Note that in all three cases, the features set $\mathcal{T}$ highlighted by the explanations are different. However, finding any feature set $\mathcal{T}$ in neural networks is challenging because features $[\mathcal{T}_i], \forall i \in [1,M]$ are not disentangled. Rather, feature extraction and inference occur simultaneously by projecting data onto network parameters. We consider $\mathcal{T}$ to be the span of network parameters or as manifolds in a high dimensional space. This manifold is, however, different for different network architectures. Hence, each feature set $\mathcal{T}$ is tied to a specific neural network $f(\cdot)$ as $\mathcal{T}^f$. To keep the notations simple, we ignore the superscript and name the three paradigms as observed correlation, observed counterfactual, and observed contrastive explanations. Observed, in this article, is from the perspective of a neural network.

\begin{figure*}[!t]
\begin{center}
\minipage{\textwidth}%
\includegraphics[width=\linewidth]{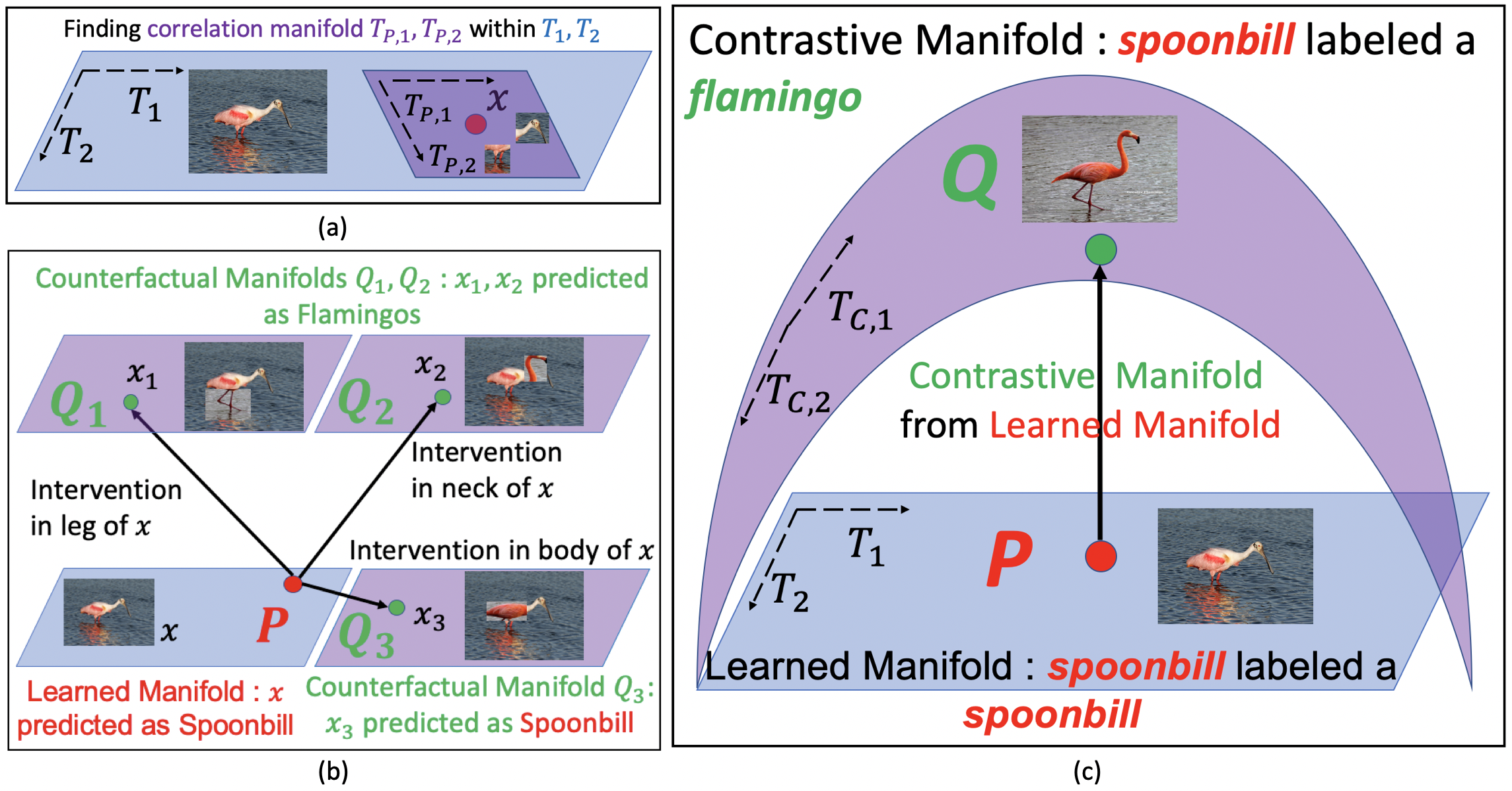}
\endminipage
\vspace{-3mm}
\caption{Three paradigms of explanations, and their manifolds. All requisite changes in the manifolds can be tracked using gradients from Eq.~\ref{eq:Backprop}.}\vspace{-0.7cm}\label{fig:Manifolds}
\end{center}
\end{figure*}

\subsection{Observed Correlation Explanations}
\label{subsec:Causal}
Observed correlation explanations are answers to \emph{`Why P?'} questions. $P$ can be any learned class but is generally the prediction made by the network. Observed correlation explanations can be modeled using any of indirect or direct explanations from Section~\ref{subsec:ExplanationMap_Background}. Using Definition~\ref{def:explanation}, observed correlation explanations $\mathcal{M}_{cu}(\cdot)$ find the feature set $\mathcal{T}_p$ that satisfy $\mathbb{P}(\mathcal{T}=\mathcal{T}_p|Y=P)$. In Fig.~\ref{fig:Manifolds}a, we visually represent the search for the observed correlation features for the example of the spoonbill image. Consider the blue manifold as the learned manifold where the image of the spoonbill is classified as a spoonbill. The set of all possible features $[\mathcal{T}_i], \forall i \in [1,M]$ exists on the blue manifold. The goal of any observed correlation explanation is to find the features $\mathcal{T}_p = \bigcup\limits_{i=1}^{K} \mathcal{T}_i$, that span the purple manifold within the larger blue manifold. While the blue manifold uses all parts of $x$ to recognize a spoonbill, only the features in its legs and neck are sufficient for $x$'s recognition. Hence $\mathcal{T}_p$ are features in a spoonbill's neck and legs.

Note that the observed correlation features presented here are different from interventionist causal factors discussed in~\cite{chalupka2014visual}\cite{scholkopf2021toward}\cite{pearl2009causal} even though the question it answers is causal. These are \emph{post-hoc} explanations observed from a network. The observed correlation explanations take the prediction $P$ into consideration to retroactively extract features $\mathcal{T}_p$. A list of common existing observed correlation techniques are shown in Table~\ref{tab:Results_Compare}.

\subsection{Observed Counterfactual Explanations}
\label{subsec:Counterfactual}
Counterfactual Explanations are contextually relevant explanations that are of the paradigm \emph{`What if?'}. The explanation to \emph{`What if the bullmastiff was not in the given image?'} is shown in Fig.~\ref{fig:Paradigms}. Other counterfactual questions can take the form \emph{`What if the dog had feathers?'} or \emph{`What if the dog had a longer neck?'}. Generally, these questions are tackled by intervening within the image in~\cite{goyal2019counterfactual}\cite{pearl2009causal}\cite{scholkopf2021toward}. In the observed paradigm, we intervene in the feature space $\mathcal{T}$. Conceptually, this is similar to maximizing $\mathbb{P}(do(\mathcal{T})|Y=y)$ from Section~\ref{subsec:Interventionist}. These \emph{do} interventions are limited since the original feature set $\mathcal{T}$ remains unchanged. Depending on the question asked, observed counterfactual explanations $\mathcal{M}_{cf}(\cdot)$, are given by the probability of $\mathcal{T}_{\hat{y}}$ that satisfy $\mathbb{P}(\mathcal{T}=\mathcal{T}_{\hat{y}}|Y=y)$. $\mathcal{T}_{\hat{y}}$ is defined similarly to observed correlation explanations except that there is an intervention in the feature set of the label depicted by $\hat{y}$. This intervention can either be the absence of $y$, like the counterfactual question in Fig.~\ref{fig:Paradigms}, or alterations in the data that changes $y$, like in~\cite{goyal2019counterfactual}. 

In Fig.~\ref{fig:Manifolds}b, we visualize three possible interventions on the spoonbill when $y = P$. The first two interventions occur on the legs and neck of the spoonbill respectively. These intervened images $x_1$ and $x_2$ are projected onto separate counterfactual manifolds and are classified as flamingos. The last intervention occurs in the body of the flamingo but is insufficient to change its classification. Nevertheless, $x_3$ is projected onto a different set of features that span a new manifold. Hence, in this toy example, there are three separate $\hat{p}$ and three $\mathcal{T}_{\hat{p}}$. Note that the underlying features, $\mathcal{T} = [\mathcal{T}_1, \mathcal{T}_2 \dots \mathcal{T}_M]$ remain the same. The union of the subsets given by $\bigcup\limits_{i=1}^{K} \mathcal{T}_i$ is different. This suggests that no new features are learned, rather a new union of existing features is formed. The result shown in Fig.~\ref{fig:Paradigms} illustrates how the network can highlight the features in the cat when denied the features of the dog. The authors in~\cite{lopez2017discovering} show that interventions can be long, complex and impractical. 

\subsection{Observed Contrastive}
\label{subsec:Contrastive}
The third explanatory paradigm of contrastive explanations provide an intuitive resolution to the intervention complexity challenge by taking advantage of the implicit notion about individual classes learned by the network. They answer questions of the form \emph{`Why P, rather than Q?'}, where $P$ is the predicted class and $Q$ is any learned contrast class. Contrastive explanations take into account the context provided by the questionnaire thus making explanations more relevant. For a given class $Q$ and predicted class P, contrastive explanations $\mathcal{M}_{ct}(\cdot)$ maximize the probability $\mathbb{P}(\mathcal{T}=\mathcal{T}_{p,q}|Y=Q)$, where $\mathcal{T}_{p,q} = \bigcup\limits_{q=1}^{K_1} \mathcal{T}_p \cap \mathcal{T}_q$. From Section~\ref{subsec:Interventionist}, we are intervening in the label space as $\mathbb{P}(\mathcal{T}|do(Y))$ to create observed contrastive paradigm.

Consider the learned blue manifold in Fig.~\ref{fig:Manifolds}c which consists of all features $\mathcal{T}$ that are used to recognize a spoonbill. Contrastive explanations require a new hypothetical manifold that produces enough interventions in $\mathcal{T}$ to predict a spoonbill as a flamingo. This occurs when the parameters of the blue manifold are changed to obtain the contrastive purple manifold. The gradients described in Eq.~\ref{eq:Backprop} is used to adjust parameters between learned and contrastive manifolds in~\cite{prabhushankar2020contrastive}.

\subsection{Probabilistic Completeness of explanations}
\label{subsec:Complete}
An advantage of defining the three observed explanatory maps and their associated paradigms in a probabilistic fashion is that they can be combined into complete explanations. The authors in~\cite{gilpin2018explaining} describe completeness in neural networks as explaining all parameters. In this article, we explain all probabilities. From Eq.~\ref{eq:Filter}, a neural network can have $N$ class probabilities. An explanatory technique must answer for any one of these possible decisions in the form of correlation-based, counterfactual, and contrastive questions. Such an explanation is termed as a complete explanation. A rigorous definition is provided below:
\begin{definition}[Complete Explanation]\label{def:Comp_Exp}
A complete explanation, $\mathcal{M}_c(\cdot)$, is a union of all functions $\mathcal{M}(\cdot)$ that simultaneously maximize the conditional probabilities $\mathbb{P}(\mathcal{T}|Y)$, for all $Y \in [1, N]$ and all features $\mathcal{T}$, i.e, $\mathcal{M}_c(\cdot) = \bigcup\limits_{j=1}^{N} \mathcal{M}_j(\cdot) = \bigcup\limits_{j=1}^{N} \mathbb{P}(\bigcup\limits_{i=1}^{K} \mathcal{T}_i|Y_j)$, 
\end{definition}
\noindent where $\mathcal{M}(\cdot), \mathcal{T}, Y$ and $K$ are defined in Definition~\ref{def:explanation}. Complete explanations can simultaneously answer questions regarding any class $Y$. For instance, in Fig.~\ref{fig:Paradigms}, given a dog and a cat image, a question of the form \emph{`Why Bluejay?'} is an acceptable question based on Definition~\ref{def:Comp_Exp} since it probes $Y = y_{\text{Bluejay}}$ and since $f(\cdot)$ has learned the Bluejay class. However, it is irrelevant seeing that there is no Bluejay in the image. An alternative and relevant question of the form \emph{`Why Bullmastiff, rather than a Bluejay?'}, can be used to probe $Y = y_{\text{Bluejay}}$. An explanation for such a question highlights the face of the dog~\cite{prabhushankar2021Contrastive}. Note that the three paradigms provide a way to probe all class probabilities in a contextually relevant fashion. 

Consider a binary classifier $f(\cdot)$ with $P$ and $Q$ as the two possible classes. Let $\mathbb{P}(P)$ and $\mathbb{P}(Q)$ be the probabilities of $f(\cdot)$ predicting a given image $x$ to belong to $P$ and $Q$, respectively.
For a binary classifier, the sum of both probabilities add to $1$, 
\begin{equation}\label{eq:Total}
    1 = \mathbb{P}(P) + \mathbb{P}(Q).
\end{equation}
Decomposing Eq.~\ref{eq:Total} using the law of total probability, we have
\begin{equation}
    1 = \mathbb{P}(P|\mathcal{T}_p)\mathbb{P}(\mathcal{T}_p) + \mathbb{P}(P|\mathcal{T}_p^c)\mathbb{P}(\mathcal{T}_p^c) + \mathbb{P}(Q|\mathcal{T}_q)\mathbb{P}(\mathcal{T}_q) + \mathbb{P}(Q|\mathcal{T}_q^c)\mathbb{P}(\mathcal{T}_q^c), 
\end{equation}
\noindent where $\mathcal{T}_p^c$ and $\mathcal{T}_q^c$ are complements. In other words, $\mathcal{T}_p^c$ refers to the set of features not used to make the decision $P$ and $\mathcal{T}_q^c$ are the features not used to decide $Q$. Using Bayes' theorem on all terms in the right hand side, by eliminating the probabilities of the features in the numerator and denominator, and rearranging to obtain $\mathcal{T}_p^c$ with $\mathcal{T}_q^c$, we get,
\begin{equation}\label{eq:all_prob}
    1 =  \mathbb{P}(\mathcal{T}_p|P)\times \mathbb{P}(P) + \mathbb{P}(\mathcal{T}_q|Q)\times \mathbb{P}(Q) + \mathbb{P}(\mathcal{T}_p^c|P)\times \mathbb{P}(P) + \mathbb{P}(\mathcal{T}_q^c|Q)\times \mathbb{P}(Q).
\end{equation}
$\mathcal{T}_p^c$ is analogous to the form \emph{`What if the bullmastiff was not in the image?'} in Fig.~\ref{fig:Paradigms}, where the network makes a decision without considering the dog features. Similarly, $\mathcal{T}_q^c$ is of the form \emph{`What if the tigercat was not in the image?'}. Hence, a function that maximizes these terms, $\mathcal{M}_{cf}(\cdot)$, are counterfactuals from Section~\ref{subsec:Counterfactual}. Therefore, any explanatory map that maximizes the probability term with $\mathcal{T}^c$ is an observed counterfactual explanation. Similarly, $\mathbb{P}(\mathcal{T}_p|P)$ and $\mathbb{P}(\mathcal{T}_q|Q)$ probability terms are maximized by functions that are observed correlation and contrastive explanations from Sections~\ref{subsec:Causal} and~\ref{subsec:Contrastive} respectively. The sum of these functions maximizes all probabilities for data $x$, and hence is a complete explanation $\mathcal{M}_c(x)$. Substituting these functions in Eq.~\ref{eq:all_prob}, and using Definition~\ref{def:Comp_Exp}, we have,
\begin{equation}\label{eq:complete}
    \mathcal{M}_c(x) =  \mathcal{M}_{cu}(x) + \mathcal{M}_{ct}(x) + \mathcal{M}_{cf}(x),
\end{equation}

Hence, a complete explanation that accounts for all probabilities in a network simultaneously answers the three observed correlation, counterfactual, and contrastive questions. In other words, an explanatory technique that answers observed correlation, counterfactual, and contrastive questions is sufficiently complete. In this article, we advocate for creating complete explanations. We show that a combination of gradient-based methods from~\cite{selvaraju2017grad}, and~\cite{prabhushankar2020contrastive} can be used to achieve such an explanation in Section~\ref{subsec:Methods}. In Section~\ref{subsec:Actual_causal}, we elaborate on the significance of a complete explanation.

\subsection{Gradient-based Complete Explanations}
\label{subsec:Methods}
In Fig.~\ref{fig:Manifolds}, the requisite change in manifolds for observed correlation, counterfactual, and contrastive explanations occurs because of the change in the network parameters that span the manifolds. From Eq.~\ref{eq:Backprop}, the change in parameters can be characterized by gradients. In this section, we show that a gradient-based technique can provide a complete explanation. The authors in~\cite{selvaraju2017grad} propose Grad-CAM, an explanatory technique that provides a gradient based localization map leading to a decision. Grad-CAM is also used to create counterfactual explanations by altering the gradients. Grad-CAM is further extended as Contrast-CAM in~\cite{prabhushankar2020contrastive} that answers contrastive questions. All these works use the same underlying explanatory mechanism. The difference is in the gradients used within each technique to access the requisite manifolds. Each of these techniques is elaborated below. 

\noindent\textbf{Grad-CAM~\cite{selvaraju2017grad}} Grad-CAM is used to visually justify the decision $P$ made by a classification network by answering \emph{`Why P?'}. The activations from the last convolutional layer of a network are used to create these visualizations since they possess high-level semantic information while retaining spatial information. For any class $i, \forall i \in [1,N]$, the logit $y_i$ is backpropagated to the feature map $A_l$ where $l$ is the last convolutional layer. The gradients at every feature map are $\frac{\partial y_i}{\partial A_l^k}$ for a channel $k$. These gradients are global average pooled to obtain importance scores $\alpha_k^{cu}$ of every feature map in $l^{th}$ layer and $k^{th}$ channel. The individual maps $A_l^k$ are multiplied by their importance scores $\alpha_k$ and averaged to obtain a heat map. The Grad-CAM map at layer $l$ for class $i$ is given by $\mathcal{M}^i = ReLU(\sum_{k=1}^K \alpha_k^{cu} A^k_l )$. 

\noindent\textbf{Counterfactual-CAM~\cite{selvaraju2017grad}} The authors in~\cite{selvaraju2017grad} extend Grad-CAM by negating the gradients as $-\frac{\partial y_i}{\partial A_l^k}$ thereby decreasing the effect of the predicted class. The resulting importance score $\alpha_k^{cf}$ de-emphasizes the region used for classification. Effectively, these regions will be eliminated and the highlighted regions are the ones that the network would have used to make its decision if the object used to make the decision were not present. Hence, it is of the \emph{`What if?'} modality.  The final counterfactual explanatory map at layer $l$ is given by $\mathcal{M} = ReLU(\sum_{k=1}^K \alpha_k^{cf} A^k_l )$.

\noindent\textbf{Contrast-CAM~\cite{prabhushankar2020contrastive}} In Grad-CAM, the importance score $\alpha_k^{cu}$ weighs the activations in a layer $l$ based on $A_l^k$'s contribution towards classification. The activations are projections on network parameters and hence have access to both the correlation and contrastive information. The goal is to obtain contrast-importance score $\alpha_k^{ct}$ that highlights contrastive information in $A_l^k$. Firstly, a loss between predicted class $P$ and any contrast class $Q$ is constructed as $J(P,Q)$. This loss function inherits the difference between the classes $P$ and $Q$. Next, $J(P,Q)$ is backpropagated within the Grad-CAM framework to obtain $\alpha_k^{ct}$ as global average pooled contrastive features i.e $\alpha_k^c = \sum_u\sum_v \nabla_{W_l} J(P,Q)$, where $u,v$ are the channel sizes at layer $l$. Contrast-CAM map for contrast class $Q$ is given by, $\mathcal{M}^i = ReLU(\sum_{k=1}^K \alpha_k^{ct} A^k_l )$. 

\begin{figure*}[!t]
\begin{center}
\minipage{\textwidth}%
\includegraphics[width=\linewidth]{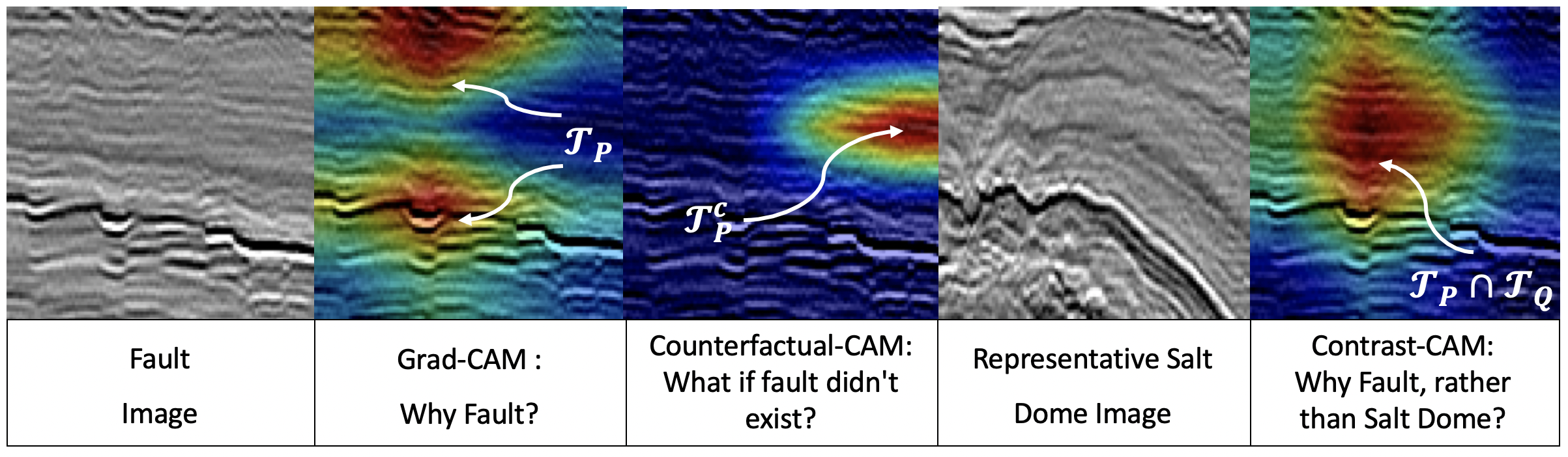}
\endminipage
\vspace{-3mm}
\caption{Observed correlation, counterfactual, representative contrast class and contrastive explanations, respectively on a fault image. The \emph{`Why Fault?'} is answered by highlighting the discontinuity features $\mathcal{T}_P$ in the rocks that correlates to presence of a fault. The counterfactual explanation analyzes the decision after removing $\mathcal{T}_P^c$. The contrastive explanation highlights features that are an intersection of $\mathcal{T}_P$ and $\mathcal{T}_Q$. Here, $P$ is the fault class and $Q$ is the salt dome class.}\vspace{-0.7cm}\label{fig:Results}
\end{center}
\end{figure*}
\subsection{Significance of Observed Explanatory Paradigms and Complete Explanations}
\label{subsec:Actual_causal}
In this section, we elaborate on the significance of all three paradigms of observed explanations.
\subsubsection{Results on non-interventionist data}
The results shown in Figs.~\ref{fig:Intro} and \ref{fig:Paradigms} are on natural images when $f(\cdot)$ is a VGG-16 architecture. The observed correlation and counterfactual explanations are extracted using~\cite{selvaraju2017grad}, and observed contrastive is extracted through Contrast-CAM~\cite{prabhushankar2020contrastive}. In Fig.~\ref{fig:Results}, a fault in the seismic section of the Netherlands F3 block~\cite{LANDMASS} is analyzed. On such data, active interventions from a causal perspective are not possible. Instead, we rely on our observed paradigms. The dataset~\cite{LANDMASS} has four classes - faults, salt domes, chaotic regions, and horizons. A ResNet-18 network is trained and the corresponding Grad-CAM~\cite{selvaraju2017grad}, Counterfactual-CAM~\cite{selvaraju2017grad}, and Contrast-CAM~\cite{prabhushankar2020contrastive} explanations extracted from the trained model are shown. The observed correlation explanation \emph{`Why fault?'}, highlights the region where faults are recognizable. The observed contrastive explanation to the question \emph{`Why fault, rather than salt dome?'}, performs the harder task of tracking the fault since the Grad-CAM explanation can also be confused to be highlighting the salt dome region. The counterfactual explanation answers \emph{`What if the fault did not exist?'}, and highlights the horizons class indicating that horizons would be the classification. All three paradigms provide new insights to seismic interpreters who are then qualified to assess the final decision. Note the complementary nature of the explanations. All features are highlighted across the seismic section, thereby visually validating Eq.~\ref{eq:complete}. 
\subsubsection{Using complete explanations to differentiate context features from observed correlation features}
The authors in~\cite{prabhushankarCausal} use the completeness of observed gradient-based methods to provide a methodology to obtain separate out context features $\mathcal{T}_c$ from observed correlation features $\mathcal{T}_p$. They analyze Grad-CAM and postulate that the Grad-CAM explanation \emph{`Why P?'} is a combination of causal and context features making it correlations. They use a set-theoretic approach to provide relationships between the two sets of features  and define context features interms of contrastive and counterfactual features. This work can be considered as an explanatory bootstrap where multiple methods are tied together to produce a new explanation. Such a bootstrap is possible because each individual method is answering a specific question.
\subsubsection{Using gradient features for applications}
\emph{Post-hoc} explanations are passive in the sense that they justify decisions once decisions are made. However, the observed contrastive and counterfactual explanations can actively be used for inference and training. The authors in~\cite{dhurandhar2018explanations} construct contrastive explanations to mine pertinent negatives to train networks. The authors in~\cite{prabhushankar2021Contrastive} use contrastive explanations and features to create robust networks. They show that usage of contrastive features $\mathcal{T}_{p,q}$ in addition to observed correlation features $\mathcal{T}_p$ increases the robustness of a network to noise by $4\%$ on \texttt{CIFAR-10C}~\cite{hendrycks2019benchmarking} dataset. This robustness is also showcased on domain shifted data where the accuracy increases by $5.2\%$ on \texttt{Office}~\cite{saenko2010adapting} dataset. The authors use these contrastive $\mathcal{T}_{p,q}$ features as a plug-in on top of existing $\mathcal{T}_p$ features thereby heuristically validating the significance of complementary observed paradigms.
\subsubsection{Analyzing explanatory techniques through observed paradigms}
By defining explanations in terms of probabilities, we tie existing explanatory techniques to reasoning and complete them based on abstract questions that use learned features $\mathcal{T}$. This provides insights into existing explanatory techniques along with a unique avenue for analyzing trends in the field of explainability. Comparisons of existing methods is performed in Table~\ref{tab:Results_Compare}. We first provide a taxonomy of evaluation for the observed paradigms before analyzing the trends. We then showcase challenges in existing explainability techniques as analyzed through the lens of our question-based paradigms.

\begin{figure*}[!htb]
\begin{center}
\minipage{\textwidth}%
\includegraphics[width=\linewidth]{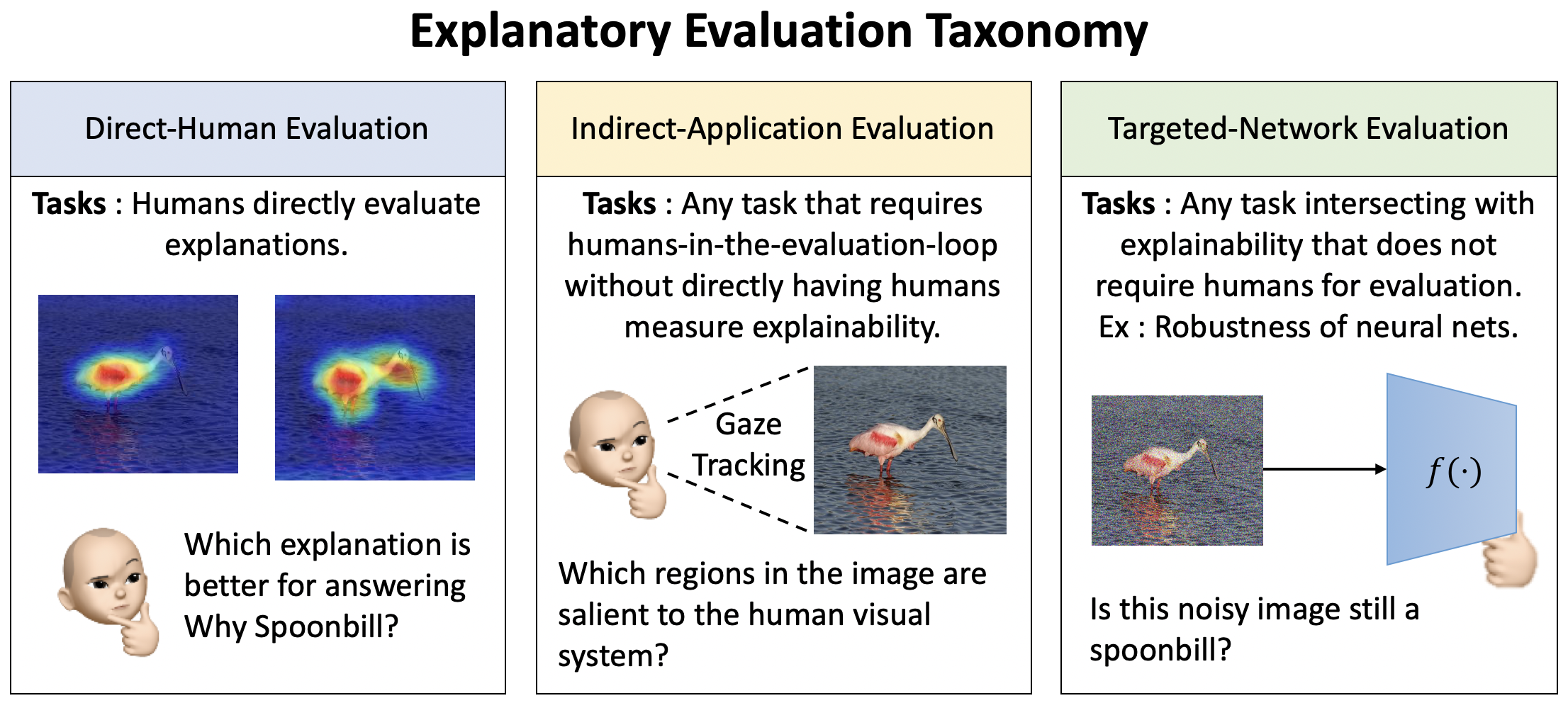}
\endminipage
\vspace{-3mm}
\caption{Three taxonomies of evaluation based on human involvement. The first is a direct evaluation of explanations by humans. The second defines explanations as proxy tasks that require human evaluation. The third uses existing tasks that have shown to require interpretability.}\vspace{-0.7cm}\label{fig:Evaluation}
\end{center}
\end{figure*}

\section{Evaluating Explanations}
\label{sec:Evaluation}
The subjectivity of explainability creates a challenge for its evaluation. For a number of applications including recognition, the only available ground-truths are object labels. Hence, explainability is not a task with ground-truth labels which can be easily evaluated. The authors in~\cite{doshi2017towards} provide a broad evaluation overview for AI techniques. We adapt this framework for neural network explainability and present it as an Explanatory Evaluation Taxonomy. An overview is provided in Fig.~\ref{fig:Evaluation}. Inherently, explanations are for humans. All evaluation criteria must be grounded in either direct or indirect interactions with humans. Consequently, there are three possible taxonomies that existing explanatory techniques use for evaluation. We first introduce and elaborate on these taxonomies. Specific explanatory paradigm techniques and their related evaluation criteria are provided in Section~\ref{sec:Existing_categories}.

\subsection{Direct-Human Evaluation}
\label{subsec:Subjective}
In this taxonomy, explanations serve to directly convince humans of the decisions made by neural networks. As such, subjectively assessing the quality of explanations involves humans in the loop. With the widespread applicability of neural networks, the most common evaluation of explainability is qualitative Direct-Human evaluation. Authors show application results that $f(\cdot)$ was trained for, like recognition accuracy, along with selected explanations for images. This is true for all data domains including natural images~\cite{zeiler2014visualizing}, medical images~\cite{prabhushankarCausal}, seismic images~\cite{shafiq2018towards} among others. Early explainability techniques including~\cite{zeiler2014visualizing, smilkov2017smoothgrad, springenberg2014striving} all follow this approach. A quantitative approach to direct-human evaluation is shown in~\cite{selvaraju2017grad}. The authors in \cite{selvaraju2017grad} create explanation maps on random $90$ images from ImageNet and use Amazon Mechanical Turk to find and ask people to rate their explanation maps as compared against explanations created by~\cite{zhou2016learning,springenberg2014striving}. Note that the goal here is not to evaluate the efficacy of the explanatory map by itself, but rather a way to compare between explanatory methods. 

\begin{figure*}[!t]
\begin{center}
\minipage{\textwidth}%
\includegraphics[width=\linewidth]{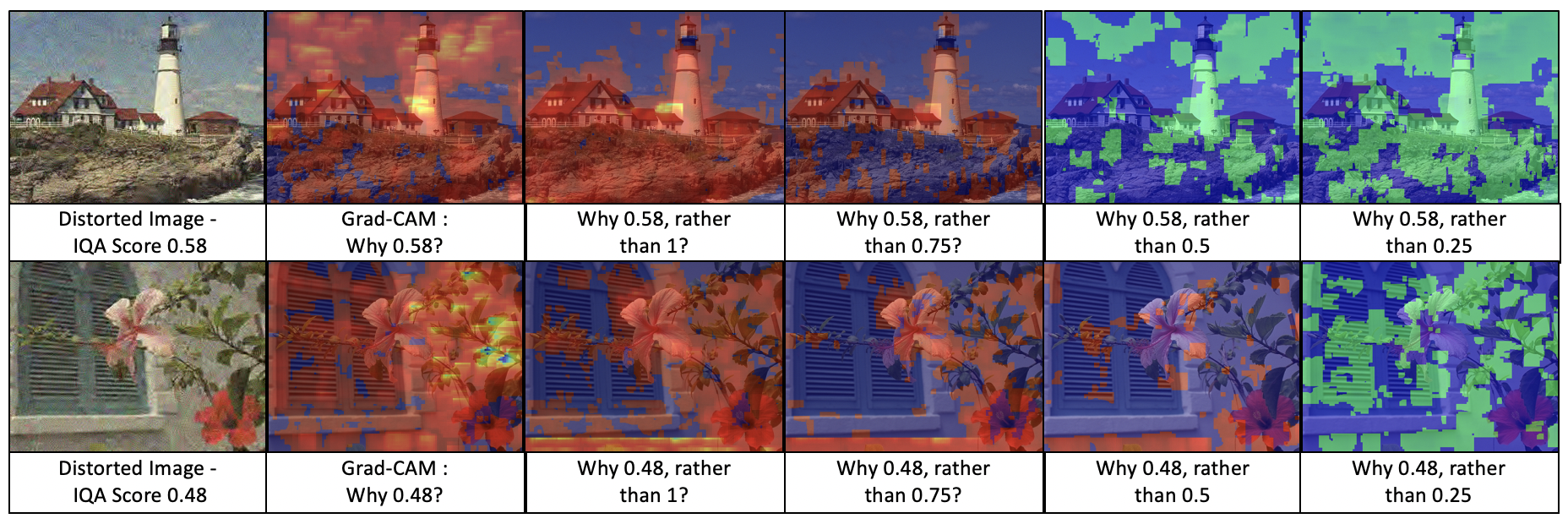}
\endminipage
\vspace{-3mm}
\caption{Grad-CAM and Contrastive explanations of questions shown below each image. Best viewed in color.}\vspace{-0.7cm}\label{fig:IQA}
\end{center}
\end{figure*}

While subjectively evaluating explanation maps is resource intensive, it provides new insights into data and applications. In the application of Image Quality Assessment (IQA), the authors in~\cite{prabhushankar2020contrastive} show that contrastive explanations provide additional context to the explanations. In Fig.~\ref{fig:IQA}, $P$ is the estimated quality score and $Q$ is a given contrast score. When $Q > P$, the contrastive explanations describe why a lower score is predicted. According to the obtained visualizations when $Q = 1$ and $Q = 0.75$, the estimated quality is primarily due to the distortions within the foreground portions of the image as opposed to the background. Further analysis is provided in~\cite{prabhushankar2020contrastive}. Since Direct-Human evaluations provide new qualitative information, the authors constructed contrastive features from existing IQA metric~\cite{temel2016unique} and showed that quantitative results increase across multiple IQA datasets~\cite{kwon2019distorted}. 

\subsection{Indirect-Application Evaluation}
\label{subsec:Objective}
In this taxonomy, tasks are designed that indirectly define explainability. Human evaluation is conducted on such tasks and since explainability is defined as a function of the task, explainability is evaluated simultaneously. The advantage of such an approach is twofold : 1) The existing framework for the application can be utilized, 2) Explanations are not directly measured which validates the original supposition that explanations are not a task by themselves but rather auxiliaries that support the reasoning behind tasks.

\noindent\textbf{Evaluation tasks: }In the the application of recognition, the authors in~\cite{zhou2016learning} define explanations as localization maps of a given object within the image. They evaluate explanations as localization, fine-grained recognition, weakly supervised localization, pattern discovery, and Visual Question Answering. The authors in~\cite{chattopadhay2018grad} derive explanation $\mathcal{M}$ for an image $x$ and mask $x$ as $x' = x \times \mathcal{M}$. They pass the masked image through the network and check for accuracy. This is conducted across a dataset of images for a number of explanatory techniques. Higher the accuracy, better is the localization capability of the explanation, and hence better is the explanation. Note that the definition of explanation here is localization of objects. The authors in~\cite{sun2020implicit} use visual saliency application as a proxy for explanation. The intuition is that human gaze lingers on the most salient parts of a given image. These salient regions are explanations for decisions made on that image since it attracts human gaze. The authors construct explanation maps using existing techniques of Grad-CAM~\cite{selvaraju2017grad} and Guided Backpropagation~\cite{springenberg2014striving} and show that their own implicit saliency tracks the human gaze with a higher correlation. Other applications include weak segmentation~\cite{selvaraju2017grad}, pointing game~\cite{petsiuk2018rise, selvaraju2017grad} and image captioning~\cite{petsiuk2018rise}.

\noindent\textbf{Evaluation metrics: }The novelty in Indirect-Application evaluation is in aligning the definition of explanations with existing tasks. Once they are aligned, the established metrics for each application is used for evaluation. For instance, classification accuracy is used to evaluate masked images in~\cite{chattopadhay2018grad}. Normalized Scanpath Saliency (NSS) and Correlation Coefficient (CC) are used in~\cite{sun2020implicit} for evaluating visual saliency.

\subsection{Targeted-Network Evaluation}

The authors in~\cite{petsiuk2018rise} question the nature of explainability as a tool for human justification and instead suggest that explanations provide an insider's view into a neural network's decision making process. Such a process need not conform to that of a human's. Explainability is connected to other applications like robustness. The goal in this taxonomy is to create experiments that increases performance on auxiliary applications thereby increasing its explainability. 

\noindent\textbf{Evaluation Tasks: }In~\cite{prabhushankar2021Contrastive}, the application of robustness is tied to explainability. The authors show observed correlation and contrastive explanations for noisy images and how having a large corpus of explanations can be robust to distortions. Consequently, the gradient features used in explanations are used to train a simple multi-layer perceptron that recognizes noisy images in a more robust fashion compared to a feed-forward network. The same robustness is observed for domain adaptation experiments. These gradients are used in other robustness applications including anomaly detection~\cite{kwon2020backpropagated}, novelty detcetion~\cite{kwon2020novelty}, and out-of-distribution detection~\cite{lee2020gradients}. Gradients used in this fashion across applications connected to explainability provide indirect validation of their explanatory capacity. The authors in~\cite{goyal2019counterfactual} propose using counterfactual explanations for the application of machine teaching - where humans learn about birds categorizations from neural networks. Note that even though humans are involved, the task for humans is not to evaluate explanations. Rather, they learn from explanations and the evaluation measures how well humans learn.    

\noindent\textbf{Evaluation Metrics: }The authors in~\cite{petsiuk2018rise} and~\cite{prabhushankarCausal} propose three new evaluation criteria - all for the application of recognition. These include Probabilistic deletion and insertion~\cite{petsiuk2018rise}, accuracy deletion and insertion~\cite{prabhushankarCausal}, and transference of features~\cite{prabhushankarCausal}. All three metrics however, are for observed correlation explanations. There are currently no specific evaluation metrics designed for observed counterfactual or contrastive explanations. The challenge is currently defining tasks for these two paradigms and the metrics follow from the tasks - similar to Indirect-Human evaluation.

\subsection{Correlation, Counterfactual, and Contrastive Evaluations}
\label{subsec:Paradigm_Evaluation}
A complete summary of existing techniques and their evaluation strategies are presented in Table~\ref{tab:Results_Compare}. These summaries provide insight into evaluation strategies followed in each of the observed correlation, counterfactual, and contrastive frameworks. Useful trends regarding evaluation, extracted from the table are presented in Section~\ref{sec:Existing_categories}.

\begin{table}[!t]
\centering
\scriptsize
\caption{Explanation Map Categorizations.}
\vspace{1mm} 
\begin{tabular}{ l c c c c c c c c c c c c c c r} 
 \toprule
    \parbox[t]{1mm}{\multirow{7}{*}{\rotatebox[origin=c]{90}{Paradigms}}} & Methods & \multicolumn{3}{c}{Definition} & \multicolumn{8}{c}{Technique Categorization} &  \multicolumn{3}{c}{Evaluation}\\ 
    
    & & \parbox[t]{4mm}{\multirow{3}{*}{\rotatebox[origin=c]{90}{Indirect}}} & 
    \parbox[t]{4mm}{\multirow{3}{*}{\rotatebox[origin=c]{90}{Direct}}} & 
    \parbox[t]{4mm}{\multirow{3}{*}{\rotatebox[origin=c]{90}{Targeted}}} &
    \parbox[t]{4mm}{\multirow{3}{*}{\rotatebox[origin=c]{90}{Implicit}}} & 
    \parbox[t]{4mm}{\multirow{3}{*}{\rotatebox[origin=c]{90}{Explicit}}} & 
    \parbox[t]{4mm}{\multirow{3}{*}{\rotatebox[origin=c]{90}{Black-Box}}} &
    \parbox[t]{4mm}{\multirow{3}{*}{\rotatebox[origin=c]{90}{White-Box}}} & 
    \parbox[t]{4mm}{\multirow{3}{*}{\rotatebox[origin=c]{90}{Intervention}}} & 
    \parbox[t]{4mm}{\multirow{3}{*}{\rotatebox[origin=c]{90}{Non-Intervention}}} &
    \parbox[t]{4mm}{\multirow{3}{*}{\rotatebox[origin=c]{90}{Gradient-based}}} & 
    \parbox[t]{4mm}{\multirow{3}{*}{\rotatebox[origin=c]{90}{Non-gradient based}}} & 
    \parbox[t]{4mm}{\multirow{3}{*}{\rotatebox[origin=c]{90}{Direct-Human}}} &
    \parbox[t]{4mm}{\multirow{3}{*}{\rotatebox[origin=c]{90}{Indirect-Application}}} & 
    \parbox[t]{4mm}{\multirow{3}{*}{\rotatebox[origin=c]{90}{Targeted-Network}}} \\
    \\
    \\
    \\
    \\
    \\
    \\
    \\
    \midrule \\
    \parbox[t]{1mm}{\multirow{18}{*}{\rotatebox[origin=c]{90}{Correlation}}} &
    Deconvolution~\cite{zeiler2014visualizing}  & \checkmark & & & & \checkmark & & \checkmark & & \checkmark & & \checkmark & \checkmark & & \\ \\
    & Inverted Representations~\cite{mahendran2015understanding} & \checkmark & & & & \checkmark & & \checkmark & & \checkmark & & \checkmark & \checkmark & & \\ \\
    & Guided-Backpropagation~\cite{springenberg2014striving} & & \checkmark & & \checkmark & & & \checkmark & & \checkmark & \checkmark & & & \checkmark & \\ \\
    & SmoothGrad~\cite{smilkov2017smoothgrad} & & \checkmark & & \checkmark & & & \checkmark & \checkmark & & \checkmark & & & \checkmark & \\ \\
    & LIME~\cite{ribeiro2016should} & & \checkmark & & & \checkmark & \checkmark & & \checkmark & & & \checkmark & \checkmark & \checkmark & \\ \\
    & CAM~\cite{zhou2016learning} & & \checkmark & & & \checkmark & & \checkmark & & \checkmark & \checkmark & & & \checkmark & \\ \\
    & Graph-CNN~\cite{zhang2018interpreting} & \checkmark & & & & \checkmark & & \checkmark & & \checkmark & & \checkmark & \checkmark & \checkmark & \\ \\
    & GradCAM~\cite{selvaraju2017grad} & & & \checkmark & \checkmark & & & \checkmark & & \checkmark & \checkmark & & \checkmark & \checkmark & \\ \\
     & TCAV~\cite{kim2018interpretability} & & \checkmark & & & \checkmark & & \checkmark & & \checkmark & \checkmark & & \checkmark & \checkmark & \\ \\
    & GradCAM++~\cite{chattopadhay2018grad} & & & \checkmark & \checkmark & & & \checkmark & & \checkmark & \checkmark & & \checkmark & \checkmark & \\ \\
    & RISE~\cite{petsiuk2018rise} & & \checkmark & & \checkmark & & \checkmark & & \checkmark & & & \checkmark & & \checkmark & \checkmark \\ \\
    & Causal-CAM~\cite{prabhushankarCausal} & & & \checkmark & \checkmark & & & \checkmark & & \checkmark & \checkmark & & \checkmark & & \checkmark \\ \\
    \midrule
    \parbox[t]{5mm}{\multirow{3}{*}{\rotatebox[origin=c]{90}{\tiny{Counterfactual}}}} &  
    Counterfactual-CAM~\cite{selvaraju2017grad} & & & \checkmark & \checkmark & & & \checkmark & & \checkmark & \checkmark & & \checkmark & \\ \\
    & Goyal et.al~\cite{goyal2019counterfactual} & & & \checkmark & \checkmark & & & \checkmark & \checkmark & & & \checkmark & \checkmark & \checkmark & \\ \\
    \midrule
    \parbox[t]{1mm}{\multirow{6}{*}{\rotatebox[origin=c]{90}{Contrastive}}} & 
    CEM~\cite{dhurandhar2018explanations} & & & \checkmark & & \checkmark & & \checkmark & \checkmark & & & \checkmark  & \checkmark & \checkmark & \\ \\
    & Contrast-CAM~\cite{prabhushankar2020contrastive} & & & \checkmark & \checkmark & & & \checkmark & & \checkmark & \checkmark & & \checkmark & & \\ \\
    & Contrastive Reasoning~\cite{prabhushankar2021Contrastive} & & & \checkmark & \checkmark & & & \checkmark & & \checkmark & \checkmark & & \checkmark & & \checkmark \\ \\
    \bottomrule
\end{tabular}
\label{tab:Results_Compare}\vspace{-3mm}
\end{table}

\section{Categorizations, Trends, and Insights into Explainability}
\label{sec:Existing_categories}
The explanatory paradigms are natural extensions of exiting works. While categorization based on abstract explanations seem intuitive, existing techniques derive their novelty by other means. Specifically, existing explanatory categorizations are based on design choices within the explanatory techniques themselves. These design choices include - need for $f(\cdot)$'s parameters during explanations, architectural design change in $f(\cdot)$ for deriving explanations among others. We first expand on these categorizations before summarizing existing methods in Table~\ref{tab:Results_Compare}.

\noindent\textbf{Implicit vs explicit explanations} A number of techniques require an architectural change to the network or a separate network to generate explanations~\cite{zeiler2014visualizing, zhou2016learning, dhurandhar2018explanations, ribeiro2016should, kim2018interpretability}. We term them as explicit explanations. In some cases, the change in network architecture can potentially alter underlying results~\cite{selvaraju2017grad}. Implicit explanations follow the original definition of explanations which is that explanations justify inference.

\noindent\textbf{Black-box vs White-box explanations} Techniques including~\cite{zeiler2014visualizing, zhou2016learning, selvaraju2017grad, prabhushankar2020contrastive} all require access to the model, its parameters and/or network states to provide explanations. These are termed white-box techniques. Other techniques exist that treat neural networks as black-boxes to make decisions~\cite{petsiuk2018rise}. Both RISE~\cite{petsiuk2018rise} and LIME~\cite{ribeiro2016should} require model outputs and not parameters.

\noindent\textbf{Interventionist vs Non-interventionist Explanations} When an explanatory technique requires changes in data, we term them as interventionist technique. SmoothGrad~\cite{smilkov2017smoothgrad} requires noise added to the images. Counterfactual visual explanations~\cite{goyal2019counterfactual} require parts of the input image substituted by another. RISE~\cite{petsiuk2018rise} requires generation of a large quantity of masked input images. Non-interventionist techniques do not require interventions in input data.

\noindent\textbf{Gradient-based vs Non-gradient based Explanations} A number of methods use gradients described in Section~\ref{subsec:Prelim} as features. Some methods like SmoothGrad~\cite{smilkov2017smoothgrad}, Guided Backpropagation~\cite{springenberg2014striving} directly use gradients as explanation maps. Others including Grad-CAM~\cite{selvaraju2017grad}, and Contrast-CAM~\cite{prabhushankar2020contrastive} use them as features within their frameworks.

The existing methods are all presented in Table~\ref{tab:Results_Compare}. They are chronologically ordered within each paradigm. The rows ascertain memberships to paradigms while there are three categorizations within columns. The first is the definitions of explanations from Section~\ref{sec:Background}, the second is the technique-based categorization presented earlier in the section, and finally the evaluation categorization from Section~\ref{sec:Evaluation}. Note that in Table~\ref{tab:Results_Compare}, a targeted definition is added to indirect and direct explanations. This includes all newer techniques related to the reasoning paradigms. The Table provides key insights into the evolution of explainability techniques and potential research directions. These include:
\begin{itemize}
    \item Correlation methods are studied more than the other explanatory paradigms. The indirect and direct definitions of explainability feeds observed correlations and has been well studied over time.  
    \item Indirect and direct explanations have given way to targeted explanations. Targeted explanations refer to explanations justifying contextually relevant questions. 
    \item Of the 10 techniques that use direct human evaluation, only three~\cite{selvaraju2017grad,chattopadhay2018grad,goyal2019counterfactual} do so by conducting random trials on humans. Others are qualitative works whose images are potentially chosen by the researchers to showcase their methods. More rigor is required in direct-human evaluation strategies.
    \item The difficulty in evaluation has encouraged authors to evaluate their works using multiple strategies in recent times. Since 2017, all but one work~\cite{prabhushankar2020contrastive} use more than one taxonomy to evaluate explanatory techniques.
    \item There is no consensus among researchers regarding the targeted-network task that evaluates explainability. The primary task using explanation masked images~\cite{chattopadhay2018grad} as well as robustness tasks~\cite{prabhushankar2021Contrastive} are used in a majority of these evaluations. 
    \item While technique-based categorization provides some novelty to the methods themselves, they are ineffective in predicting the trends in explainability over time.
\end{itemize}
\begin{figure*}[!t]
\begin{center}
\minipage{\textwidth}%
\includegraphics[width=\linewidth]{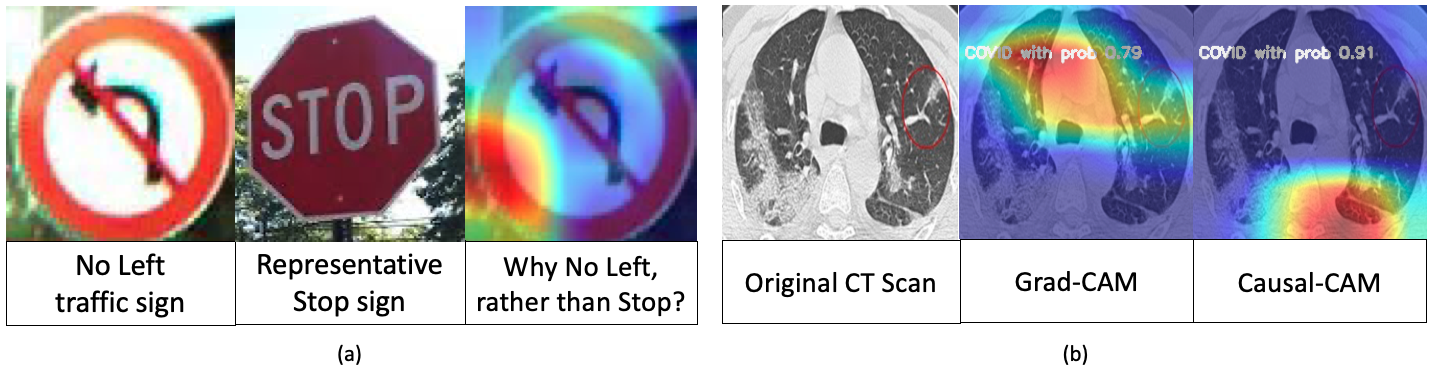}
\endminipage
\vspace{-3mm}
\caption{The contrastive explanation in (a) shows the disparity between human and machine explanations. Both methods in (b) fail to highlight the red circle.}\vspace{-0.7cm}\label{fig:Challenges}
\end{center}
\end{figure*}

\section{Challenges in Explainability and Future Research Directions}
\label{sec:Challenges}

\subsection{Disparity between human and machine explanations}
Consider the case in Fig.~\ref{fig:Challenges}a when a network, trained on the CURE-TSR~\cite{temel2017cure} traffic sign dataset, is analyzed constructively. Given a No-Right sign and asked a contrastive question of the form \emph{`Why No-Left, rather than Stop?'}, the network highlights the bottom left corner of the image. The letters that spell STOP not being in the sign is the intuitive human response to the above question. This disparity between human intuition and machine explanation is explained based on the dataset. Among the 14 traffic signs in CURE-TSR, Stop sign is the only class that has a hexagonal shape. Hence, the network has learned to check for a straight side in the bottom left. The absence of this side in $x$ indicates to the network that $x$ is not a STOP sign. Hence, the observed contrastive explanation clearly illustrates the disparity between the notion of classes between humans and machines while observed correlations could not. This calls for the need for considering complete explanations in the absence of interventions. It is not very clear \emph{if} the two explanations must align in all applications. However, for safety-critical applications in medical fields, it is required that these anomalies are analyzed and understood.

\subsection{Failure of existing objective evaluation metrics}
In Fig.~\ref{fig:Challenges}b, we visualize Grad-CAM~\cite{selvaraju2017grad} and Causal-CAM~\cite{prabhushankarCausal} maps. The original scan is from a COVID positive patient. Both the Grad-CAM and Causal-CAM explanations fail to highlight the circled red region that depicts COVID. Feeding the masked image back into $f(\cdot)$ for indirect-application evaluation, the network classifies both correctly but with a higher confidence for Causal-CAM. Hence, in both the Probabilistic and Accuracy-Deletion and Insertion metrics, Causal-CAM performs better than Grad-CAM. This essentially shows that the network has learned to classify correctly but with the wrong features. Such a feature discrepancy is not reflected in the objective metrics of evaluation. In real-world biomedical applications like in COVID-19 detection, it is imperative to identify and make decisions that are interpretable. Further study is required in designing better objective metrics and networks whose features are more human interpretable. 

\section{Conclusion}
\label{sec:Conclusion}
The key takeaways from this article include : ($i$) realizing explanations as reasoning paradigms, ($ii$) definition of complete explanations in neural networks based on reasoning ($iii$) utility of gradient-based explanatory paradigm's replicability across multifarious applications.

\subsection{Explanations as Reasoning Paradigms}
By having explanations answer targeted questions, the goal of explainability shifts from justifying decisions to reviewing and participating in making decisions. This is inline with abductive reasoning scheme that humans often use in practice. 

\subsection{Complete Explanations}
Existing techniques treat each correlation, counterfactual, and contrastive paradigms as separate entities. However, each paradigm answers complementary questions. A complete explanation is one that can answer any given question. In practice, the combination of Grad-CAM~\cite{selvaraju2017grad}, Counterfactual-CAM~\cite{selvaraju2017grad}, and Contrast-CAM~\cite{prabhushankar2020contrastive} achieve this. Similar research into methodologies that unify these paradigms and provide complete explanations is required and pertinent.

\subsection{Applicability of gradients-as-features}
From Table~\ref{tab:Results_Compare}, it is clear that gradients are used as features in explanations across multiple categorizations and paradigms. The same gradients are also used as features for recognition robustness~\cite{prabhushankar2021Contrastive}, anomaly~\cite{kwon2020backpropagated}, novelty~\cite{kwon2020novelty} and out-of-distribution detection robustness~\cite{lee2020gradients}, Image Quality Assessment~\cite{kwon2019distorted} and Visual Saliency Detection~\cite{sun2020implicit}. The adoption of a network's gradients as features across applications is a sign of its multifarious utility.

\bibliographystyle{IEEEbib}
\bibliography{references}

\end{document}